\documentclass[runningheads]{llncs}
\usepackage{graphicx}
\usepackage{amsmath}
\begin{document}
\title{Can We Spot the ``Fake News''\\ Before It Was Even Written?}
%
%
\author{
Preslav Nakov\inst{1}\orcidID{0000-0002-3600-1510}
}
\authorrunning{P. Nakov}

\institute{
Qatar Computing Research Institute, HBKU, Qatar\\ \email{pnakov@hbku.edu.qa}
}

\maketitle              

\begin{abstract}
Given the recent proliferation of disinformation online, there has been also growing research interest in automatically debunking rumors, false claims, and ``fake news.'' A number of fact-checking initiatives have been launched so far, both manual and automatic, but the whole enterprise remains in a state of crisis: by the time a claim is finally fact-checked, it could have reached millions of users, and the harm caused could hardly be undone. An arguably more promising direction is to focus on fact-checking entire news outlets, which can be done in advance. Then, we could fact-check the news before it was even written: by checking how trustworthy the outlets that published it is. We describe how we do this in the Tanbih news aggregator, which makes readers aware of what they are reading. In particular, we develop media profiles that show the general factuality of reporting, the degree of propagandistic content, hyper-partisanship, leading political ideology, general frame of reporting, and stance with respect to various claims and topics.

\end{abstract}

\section{The ``Fake News'' Phenomenon}

Recent years have seen the rise of social media, which have enabled people to easily share information with a large number of online users, without quality control.
On the bright side, this has given the opportunity to anybody to become a content creator, and it has enabled a much faster information dissemination.
On the not-so-bright side, it has also made it easy for malicious actors to spread disinformation much faster, potentially reaching very large audiences.
In some cases, this included building sophisticated profiles for individual users based on a combination of psychological characteristics, metadata, demographics, and location, and then micro-targeting them with personalized ``fake news'' and propaganda campaigns that have been weaponized with the aim of achieving political or financial gains.

To be clear, false information in the news has always been around, e.g.,~think of tabloids. However, social media have changed everything. They have made it possible for malicious actors to micro-target specific demographics, and to spread disinformation much faster and at scale, at the disguise of news. Thanks to social media, the news could be weaponized at an unprecedented scale.

As social media are optimized for user engagement, ``fake news'' thrive on these platforms, as users cannot recognize it and thus they share it, which can be further amplified by bots.
Studies have shown that 70\% of the users cannot distinguish real from ``fake news'', and that ``fake news'' spread in social media six times faster than real ones~\cite{Vosoughi1146}.
``Fake news'' are like spam on steroids: if a spam message reaches a thousand people, it would die there; in contrast, ``fake news'' can be shared again and again, and it could eventually reach millions.

As a result, there have been growing concerns about disinformation in recent years. The year 2016 gave rise to the term \textit{fake news}, as the general public got concerned about the dangers of possible manipulations of major events such as the 2016 US Presidential elections and Brexit. Then along came year 2020, when a new term was coined, \textit{infodemic}, which reflects the general concern about the spread of disinformation related to the COVID-19 pandemic. Yet, the real underlying problem all along has been the \textit{weaponization} of the news.

In the public discourse, \emph{fake news} is the preferred term when discussing the issue. Even though declared Word of the Year 2016 by Macquarie Dictionary and of Year 2017 by Collins dictionary, we find the term unhelpful, as it can easily mislead people to only focus on the veracity aspect.
At the EU level, a more precise term is preferred: \textit{disinformation},\footnote{\url{http://eeas.europa.eu/topics/countering-disinformation\_en}} 
which refers to information that is both (\emph{i})~\textit{false}, and (\emph{ii})~\textit{intents to harm}. The often-ignored latter aspect is the real reason why the society started worrying back in 2016.

Another problem with the term \textit{fake news} is that it has no generally agreed-upon definition. For example, the Meriam-Webster dictionary finds the term ``self-explanatory'' and sees no need to include it in their dictionary,\footnote{\url{http://www.merriam-webster.com/words-at-play/the-real-story-of-fake-news}} arguing that it was not a new term and that it was already in use back in the 1890s. However, ``fake news'' is an oxymoron since, as news is supposed to be true.
The Collins dictionary defines the term as ``\textit{false, often sensational, information disseminated under the guise of news reporting}'',\footnote{\url{http://www.collinsdictionary.com/dictionary/english/fake-news}} and it does not mention the intention to do harm, and the use of social media and micro-targeting. The Cambridge dictionary does better and defines it as ``\textit{false stories that appear to be news, spread on the internet or using other media, usually created to influence political views or as a joke}'',\footnote{\url{http://dictionary.cambridge.org/dictionary/english/fake-news}} but it does not mention social media and micro-targeting, and it covers satire, which is protected free speech.\footnote{\url{http://en.wikipedia.org/wiki/Hustler\_Magazine\_v.\_Falwell}}

Over time, the term \textit{fake news} started meaning different things to different people, and for some politicians, even ``news that I do not like.'' Yet, the fundamental issue with the term is that it misleads people to focus on veracity and to ignore the intention to do harm.
Focusing on the veracity aspect only is dangerous as weaponized news do not need to lie. For instance, it could cherry-pick the facts, it could appeal to emotions when presenting actual events, and so on.
In essence, it could tell the truth, only the truth, but not the whole truth.

Overall, thanks to social media, people today are much more likely to believe in conspiracy theories. For example, according to a 2019 study, 57\% of Russians believed that the USA did not put a man on the Moon. In contrast, when the event actually occurred, there was absolutely no doubt about it in the USSR, and Neil Armstrong was even invited to visit Moscow, which he did.

Indeed, disinformation has become a global phenomenon: a number of countries had election-related issues with ``fake news''. To get an idea of the scale, 150 million users on Facebook and Instagram saw inflammatory political ads, and Cambridge Analytica had access to the data of 87 million Facebook users in the USA, which it used for targeted political advertisement; for comparison, the 2016 US Presidential elections in the USA were decided by 80,000 voters in three key states.

While initially the focus has been on influencing the outcome of political elections, ``fake news'' has also caused direct life loss.
For example, disinformation on WhatsApp has resulted in people being killed in India, and disinformation on Facebook was responsible for the Rohingya genocide, according to a UN report.
Disinformation can also put people's health in danger, e.g.,~think of the anti-vaccine websites and the damage they cause to public health worldwide, or of the ongoing COVID-19 pandemic, which has also given rise to the first global infodemic.

\section{Related Work}

Recently, there has been a lot of research interest in studying disinformation and bias in the news and in social media. 
This includes challenging the truthiness of claims~\cite{Atanasova:2019:AFU:3331015.3297722,mihaylov-nakov:2016:P16-2,EMNLP2019:fauxtography}, of news~\cite{brill2001online,finberg2002digital,credibleNews,Hardalov2016,RANLP2017:clickbait,CIKM2020:FANG,Potthast2018}, of news sources \cite{D18-1389}, of social media users~\cite{CoNLL2019:troll:roles,SeminarUsers2017,Mihaylov2015FindingOM,Mihaylov2015ExposingPO,InternetResearchJournal:2018,RANLP2017:credibility:trolls}, and of social media~\cite{Canini:2011,Castillo:2011:ICT:1963405.1963500,shaar-etal-2020-known,PlosONE:2016}, as well as studying credibility, influence, and bias~\cite{Ba:2016:VERA,D18-1389,Chen:2013:BIW:2492517.2492637,Kulkarni:2018:EMNLP,Mihaylov2015FindingOM,InternetResearchJournal:2018}.
The interested reader can also check several recent surveys that offer a general overview on ``fake news''~\cite{Lazer1094}, or focus on topics such as the process of proliferation of true and false news online \cite{Vosoughi1146}, on fact-checking \cite{thorne-vlachos:2018:C18-1}, on data mining \cite{Shu:2017:FND:3137597.3137600}, or on truth discovery in general \cite{Li:2016:STD:2897350.2897352}.
For some specific topics, research was facilitated by specialized shared tasks such as the SemEval-2017 task~8 and the SemEval-2019 task~7 on Determining Rumour Veracity and Support for Rumours (RumourEval) \cite{derczynski-EtAl:2017:SemEval,gorrell-etal-2019-semeval}, the CLEF 2018--2020 CheckThat! lab on Automatic Identification and Verification of Claims \cite{clef2018checkthat:task1,clef-checkthat-T1:2019,CheckThat:ECIR2020,clef2018checkthat:task2,clef-checkthat:2020,CheckThat:ECIR2019,clef-checkthat:2019,clef-checkthat-ar:2020,clef-checkthat-T2:2019,clef2018checkthat:overall,clef2018checkthat,clef-checkthat-en:2020}, the FEVER-2018 and FEVER-2019 tasks on Fact Extraction and VERification \cite{thorne-EtAl:2018:N18-1,thorne-etal-2019-fever2}, and the SemEval-2019 Task 8 on Fact Checking in Community Question Answering Forums \cite{SemEval2019-task8,AAAI2018:factchecking}, among others.

Finally, note that the veracity of information is a much bigger problem than just ``fake news''. It has been suggested that ``Veracity'' should be seen as the fourth ``V'' of Big Data, along with Volume, Variety, and Velocity.\footnote{\url{http://www.ibmbigdatahub.com/sites/default/files/infographic\_file/4-Vs-of-big-data.jpg}}

\section{Focusing on the Source}

In order to fact-check a news \emph{article}, we can analyze its contents, e.g.,~the language it uses, and the reliability of its source, which can be represented as a number between 0 and 1, where 1 indicates a very reliable source, and 0 stands for a very unreliable one:

\begin{align}
\label{eq:article}
    factuality\left(article\right) & = reliability\left(language\left(article\right)\right) \nonumber \\
    & \qquad + reliability\left(website\left(article\right)\right)
\end{align}

In order to fact-check a \emph{claim} (as opposed to an \emph{article}), we can retrieve articles discussing the claim, then we can detect the stance of each article with respect to the claim, and we can take a weighted sum (here, the stance is a number between -1 and 1, where it is -1 if the article disagrees with the claim, it is 1 if it agrees, and it is 0 if it just discusses the claim or if it is unrelated):

\begin{equation}
\label{eq:claim}
    factuality(claim) = \sum_i \left[reliability\left(article_i\right) * stance\left(article_i, claim\right)\right]
\end{equation}

Note that in formula~(\ref{eq:article}), the reliability of the website that hosts an article serves as a prior to compute the factuality of an article, while in formula~(\ref{eq:claim}), we use the factuality of the retrieved articles to compute a factuality score for a target claim. The idea is that if a reliable article agrees/disagrees with the claim, this is a good indicator for it being true/false, and it is the other way around for unreliable articles.

Of course, the formulas above are oversimplifications, e.g.,~one can fact-check a claim based on the reactions of users in social media \cite{10.1145/3274351}, based on the claim's spread over time in social media \cite{ma2016detecting}, based on information in a knowledge graph \cite{10.1007/978-3-030-30796-7_20}, extracted from the Web \cite{RANLP2017:factchecking:external} or from Wikipedia \cite{thorne-EtAl:2018:N18-1}, using similarity to previously fact-checked claims \cite{shaar-etal-2020-known}, etc. Yet, the formulas give the general idea that the reliability of the source should be an important element of fact-checking articles and claims. Yet, it is an understudied problem.

Characterizing entire news outlets is an important task on its own right. We argue that it is more useful than fact-checking claims or articles, as it is hardly feasible to fact-check every single piece of news. Doing so also takes time, both to human users and to automatic programs, as they need to monitor the way reliable media report about a given target claim, how users react to it in social media, etc., and it takes time to get enough such evidence accumulated in order to be able to make a reliable prediction. It is much more feasible to check entire news outlets. Note that we can also fact-check a number of sources in advance, and we can then \textbf{fact-check the news before it was even written!} Once it is published online, it would be enough to check how trustworthy the outlets that published it are, in order to get an initial (imperfect) idea about how much we should trust this news. This would be similar to the movie \emph{Minority Report}, where the authorities could detect a crime before it was even committed.

In general, fighting disinformation is not easy; as in the case of spam, this is an adversarial problem, where the malicious actors constantly change and improve their strategies. Yet, when they share news in social media, they typically post a link to an article that is hosted on some website. This is what we are exploiting: we try to characterize the news outlet where the article is hosted. This is also what journalists typically do: they first check the source. 

Finally, even though we focus on the source, our work is also compatible with fact-checking a claim or a news article, as we can provide an important prior and thus help both algorithms and human fact-checkers that try to fact-check a particular news article or a claim.

How can we profile a news source? Note that disinformation typically focuses on emotions, and political propaganda often discusses moral categories \cite{IJCAI2020:propaganda:survey}. There are many incentives for news outlets to publish articles that appeal to emotions: (\emph{i})~this has a strong propagandistic effect on the target user, (\emph{ii})~it makes it more likely to be shared further by the users, and (\emph{iii})~it will be favored as a candidate to be shown in other users' newsfeed as this is what algorithms on social media optimize for. And news outlets want to get users to share links to their content in social media as this allows them to reach larger audience. This kind of language also makes them potentially detectable for Artificial Intelligence (AI) systems; yet, these outlets cannot do much about it as changing the language would make their message less effective and it would also limit its spread.

While the analysis of the language used by the target news outlet is the most important information source, we can also consider information in Wikipedia and in social media, traffic statistics, and the structure of the target site’s URL as shown in Figure~\ref{fig:evaluation_sources}:

\begin{figure}
    \centering
   \includegraphics[width=\textwidth]{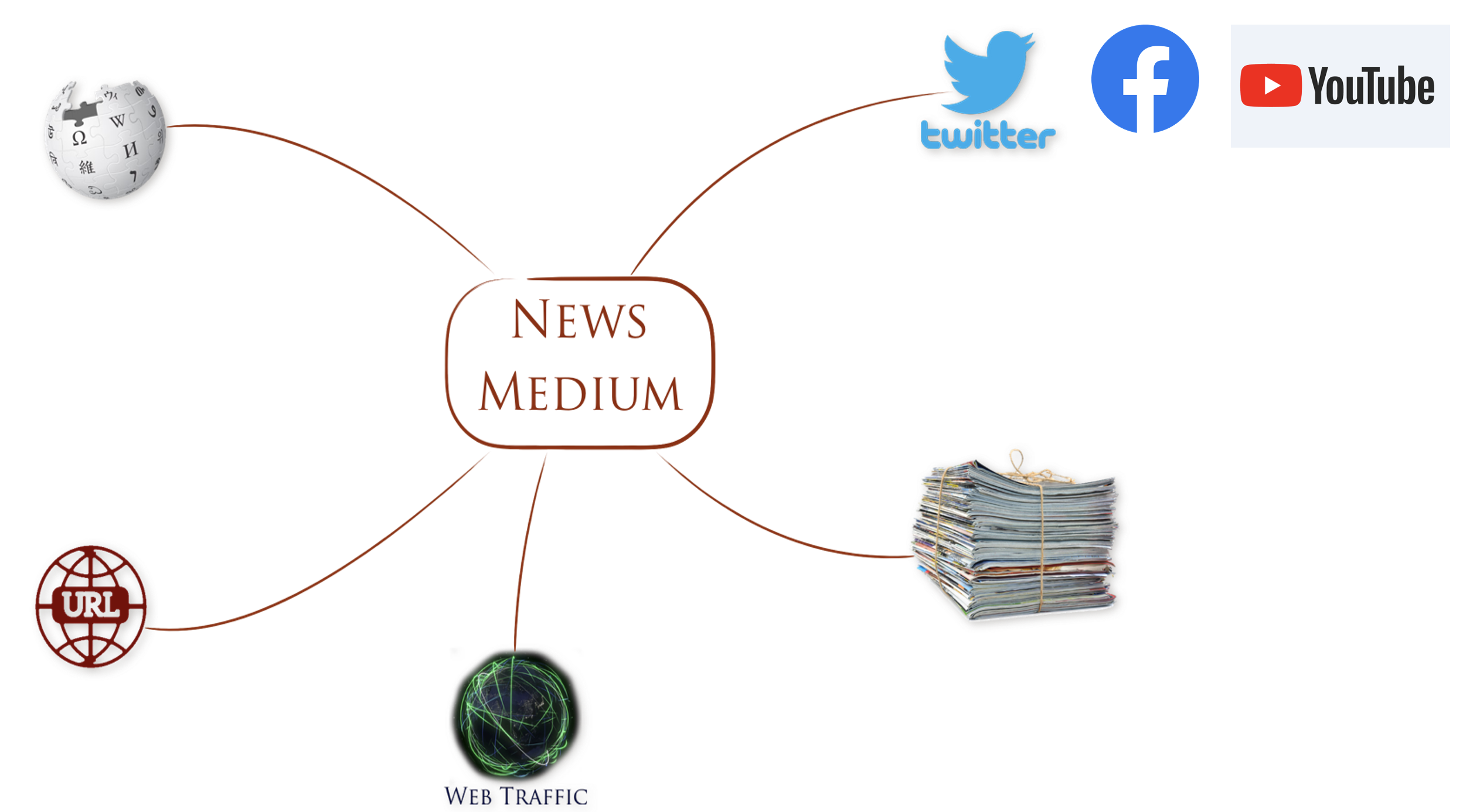}
    \caption{Information sources used to predict the factuality of reporting of a news outlet.}
    \label{fig:evaluation_sources}
\end{figure}

\begin{enumerate}
\item the text of a few hundred articles published by the target news outlet, analyzing the style, subjectivity, sentiment, offensiveness \cite{OffenseEval:NAACL:2019,OffenseEval:SemEval:2019,zampieri-etal-2020-semeval,SOLID}, toxicity~\cite{RANLP2019:toxicity}, morality, vocabulary richness, propagandistic content, etc.;
\item the text of its Wikipedia page (if any), including infobox, summary, content, categories, e.g., it might say that the website spreads false information and conspiracy theories;
\item metadata and statistics about its Twitter account (if any): is it an old account, is it verified, is it popular, how is the medium self-describing, is there a link to its website, etc.;
\item whether people in social media, e.g., on Twitter, post links in articles to the target sources in the context of a polarizing topic, and which side of the debate are these users from;
\item whether there is liberal-vs-moderate-vs-conservative bias of the audience of the target medium in social media, e.g.,~in Facebook;
\item the language used in videos by the target medium, e.g.,~in their Youtube channels (if any), where the focus is on analysis of the speech signal, i.e.,~not on what is said but on how it is said, e.g.,~is it emotional \cite{ASRU2019:deception};
\item Web traffic information: whether this is a popular website;
\item the structure of site's URL: is it too long, does it contain a sequence of meaningful words, does it have a fishy suffix such as ``.com.co'', etc.
\end{enumerate}



\section{The Tanbih Mega-Project}

Characterizing media in terms of factuality of reporting and bias is part of a larger effort at the Qatar Computing Research Institute, HBKU: the Tanbih mega-project\footnote{\url{http://tanbih.qcri.org/}} aims to limit the effect of ``fake news'', disinformation, propaganda and media bias by making users aware what they are reading, thus promoting media literacy and critical thinking.

The mega-project’s flagship initiative is the Tanbih news aggregator~\cite{EMNLP2019:tanbih},\footnote{\url{http://www.tanbih.org/}} which shows real-time news from a variety of news sources~\cite{Staykovski:19}. It builds a profile for each news outlet, showing a prediction about the factuality of its reporting\cite{D18-1389,baly-etal-2020-written,source:multitask:NAACL:2019}, its leading political ideology \cite{INTERSPEECH2019:youtube}, degree of propaganda~\cite{Barron:19,AAAI2019:proppy}, hyper-partisanship~\cite{saleh-etal-2019-team,shaprin-etal-2019-team}, and general frame of reporting (e.g., political, economic, legal, cultural identity, quality of life, etc.), and stance with respect to various claims and topics~\cite{baly-EtAl:2018:N18-2,ICWSM2020:Unsupervised:Stance:Twitter,NAACL2018:stance,EMNLP2019:Stance:crosslanguage:contrastive,stefanov-etal-2020-predicting}. For the individual news, it signals when an article is likely to be propagandistic. It further mixes Arabic and English news, and allows the user to see them all in English/Arabic thanks to QCRI’s Machine Translation technology. Tanbih also offers analytics capabilities, allowing a user to explore the media coverage, the frame of reporting, and the propaganda around topics such as Brexit, Sri Lanka bombings, and COVID-19.\footnote{\url{http://www.tanbih.org/subject/4535/CORONAVIRUS\%20OUTBREAK\%2019-20}} Moreover, it performs fine-grained analysis of the propaganda techniques in the news \cite{DaSanMartinoSemeval20task11,NLP4IF2019:propaganda:task,da-san-martino-etal-2020-prta,EMNLP2019:propaganda:finegrained,NeurIPS2019:propaganda}.\footnote{\url{http://www.tanbih.org/prta}}

We developed tools such as a Web browser plugin,\footnote{\url{http://chrome.google.com/webstore/detail/tanbih/igcppjdbignhkiikejdjpjemejoognen}} a mechanism to share media profiles and stories in social media, a tool to detect check-worthiness for English and Arabic~\cite{RANLP2017:debates,NAACL2018:claimrank,RANLP2019:checkworthiness:multitask},\footnote{\url{http://claimrank.qcri.org/}} a Twitter fact-checking bot,\footnote{\url{http://twitter.com/factchecker\_bot/}} and an API to the Tanbih functionality.\footnote{\url{http://app.swaggerhub.com/apis/yifan2019/Tanbih/0.6.0\#/}} The latter is used by Aljazeera and other partners. More recently, we have been developing tools for fighting the {COVID}-19 infodemic by modeling the perspective of journalists, fact-checkers, social media platforms, policy makers, and the society~\cite{alam2020call2arms,alam2020fighting}. 

Tanbih was developed in close collaboration with MIT-CSAIL.\footnote{\url{http://qcri.csail.mit.edu/node/25}} We were also partners in a large NSF project on Credible Open Knowledge Networks,\footnote{\url{http://cokn.org/}} and we further collaborate with Carnegie Mellon University in Qatar, Qatar University, Sofia University, the University of Bologna, Aljazeera, Facebook, the United Nations, Data Science Society, and A Data Pro, among others. As part of a larger team, including Al Jazeera, Associated Press, RTE Ireland, Tech Mahindra, and Metaliquid, and V-Nova, we won an award at IBC 2019 by TM Forum and IBC 2019 for our Media-Telecom Catalyst project on AI Indexing for Regulatory Practice.\footnote{\url{http://www.tmforum.org/ai-indexing-regulatory-practise/}}

In its 2.5 years of history, the Tanbih mega-project has produced 30+ top-tier publications, a Best Demo Award (honorable mention) at ACL-2020, and several patent applications. The project was featured in 30+ keynote talks, and it was highlighted by 100+ media including Forbes, Boston Globe, Aljazeera, MIT Technology Review, Science Daily, Popular Science, Fast Company, The Register, WIRED, and Engadget.

\section{The Future of ``Fake News''}

It is widely believed that ``fake news'' can and has affected major political events. In reality, the true impact is unknown; however, given the buzz that was created, we should expect a large number of state and non-state actors to give it a try.

From a technological perspective, we ca expect further advances in ``deep fakes'', such as machine-generated videos, and images.
This is a really scary development, but probably only in the mid-long run; as of present, ``deep fakes'' are still easy to detect both using AI and also by experienced users.

We also expect advances in automatic news generation, thanks to recent developments such as GPT-3. This is already a reality and a sizable part of the news we are consuming daily are machine generated, e.g., about the weather, the markets, and sport events. Such software can describe a sport event from various perspectives: neutrally or taking the side of the winning or the losing team. It is easy to see how this can be used for disinformation purposes.

Yet, we hope to see ``fake news'' gone the way of spam: not entirely eliminated (as this is impossible), but put under control.
AI has already helped a lot in the fight against spam, and we expect that it would play a key role in putting ``fake news'' under control as well.

A key element of the solution would be limiting the spread.
Social media platforms are best positioned to do this on their own platforms. Twitter has suspended more than 70 million accounts in May and June 2018, and these efforts continue to date; this can help in the fight against bots and botnets, which are the new link farms: 20\% of the tweets during the 2016 US Presidential campaign were shared by bots. Facebook, from its part, warns users when they try to share a news article that has been fact-checked and identified as fake by at least two trusted fact-checking organizations, and it also downgrades ``fake news'' in its news feed. We expect the AI tools used for this to get better, just like spam filters have improved over time.

Yet, the most important element of the fight against disinformation is raising user awareness and develop critical thinking. This would help limit the spread as users would be less likely to share it further. We believe that practical tools such as the ones we develop in the Tanbih mega-project would help in that respect.

\bibliography{references.bib}
\bibliographystyle{splncs04.bst}

\end{document}